\documentclass{article}

\usepackage{microtype}
\usepackage{graphicx}
\usepackage{xcolor}
\usepackage{booktabs} 

\usepackage{hyperref}

\usepackage[preprint]{icml2025}

\usepackage{amsmath}
\usepackage{amssymb}
\usepackage{mathtools}
\usepackage{amsthm}

\usepackage{tikz}
\usetikzlibrary{decorations.pathmorphing, decorations.pathreplacing, positioning, arrows.meta}

\usepackage[capitalize,noabbrev]{cleveref}
\creflabelformat{equation}{#2\textup{#1}#3}

\usepackage{subcaption}

\theoremstyle{plain}
\newtheorem{theorem}{Theorem}[section]
\newtheorem{proposition}[theorem]{Proposition}

\theoremstyle{definition}
\newtheorem{definition}[theorem]{Definition}

\newtheorem{problem}[theorem]{Problem}

\theoremstyle{remark}
\newtheorem{remark}[theorem]{Remark}

\usepackage[textsize=tiny]{todonotes}

\newcommand{\R}{{\mathbb R}}
\newcommand{\N}{{\mathbb N}}

\newcommand{\h}{\ell}
\newcommand{\Rh}{{\mathbb R}^{\ell}}

\newcommand{\normalchar}{{\mathcal N}}
\DeclareMathOperator*{\E}{\mathbb{E}}

\newcommand{\ones}[1]{{\mathbf{1}_{#1}}}
\newcommand{\zeros}[1]{{\mathbf{0}_{#1}}}
\newcommand{\eye}[1]{{\mathbf{I}_{#1}}}

\newcommand{\defeq}{\vcentcolon=}

\DeclareMathOperator*{\argmin}{arg\,min}

\newcommand{\isep}{\mathrel{{.}\,{.}}\nobreak}

\newcommand{\floor}[1]{\left\lfloor #1 \right\rfloor}

\newcommand{\bx}{\mathbf{x}}
\newcommand{\by}{\mathbf{y}}
\newcommand{\bz}{\mathbf{z}}
\newcommand{\bu}{\mathbf{u}}
\newcommand{\bv}{\mathbf{v}}

\newcommand{\Sc}{\mathcal{S}}

\newcommand{\F}{\mathcal{F}}
\newcommand{\Fseq}{\mathcal{F}_{\Rh \to \Rh}}

\newcommand{\inlinesubsection}[1]{\noindent\textbf{#1.}\ }
\newcommand{\theoremspace}{\vspace{0.2ex}}

\icmltitlerunning{Spooky Action at a Distance}

\begin{document}

\twocolumn[
\icmltitle{Spooky Action at a Distance: Normalization Layers Enable\\ Side-Channel Spatial Communication}

\icmlsetsymbol{equal}{*}

\begin{icmlauthorlist}
\icmlauthor{Samuel Pfrommer}{berkeley}
\icmlauthor{George Ma}{berkeley}
\icmlauthor{Yixiao Huang}{berkeley}
\icmlauthor{Somayeh Sojoudi}{berkeley}
\end{icmlauthorlist}

\icmlcorrespondingauthor{Samuel Pfrommer}{sam.pfrommer@berkeley.edu}

\icmlaffiliation{berkeley}{University of California, Berkeley}

\icmlkeywords{Machine Learning, ICML}

\vskip 0.3in
]



\printAffiliationsAndNotice{}  

\begin{abstract}
    This work shows that normalization layers can facilitate a surprising degree of communication across the spatial dimensions of an input tensor.
    We study a toy localization task with a convolutional architecture and show that normalization layers enable an iterative message passing procedure, allowing information aggregation from well outside the local receptive field.
    Our results suggest that normalization layers should be employed with caution in applications such as diffusion-based trajectory generation, where maintaining a spatially limited receptive field is crucial.
\end{abstract}



\section{Introduction}\label{sec:intro}
Normalization layers are widely used in deep learning architectures. They fundamentally act to shift intermediate network activations to have zero mean and unit variance.
Many normalization variants have been proposed, differing in which dimensions they normalize over; we specifically highlight BatchNorm \citep{ioffe2015batch}, LayerNorm \citep{ba2016layer}, GroupNorm \citep{wu2018group}, and InstanceNorm \citep{ulyanov2016instance}, with formal definitions deferred to \Cref{sec: normalization_layers}.
With the notable exception of LayerNorm in transformers \citep{vaswani2017attention}, these layers all generally normalize over spatiotemporal input dimensions (e.g., height and width in images).

While generally effective, normalization layers have been found to occasionally induce failures.
In Generative Adversarial Networks (GANs), normalization layers have been connected to blob-shaped output artifacts \citep{karras2020analyzing,karras2024analyzing}.
For small batch sizes, BatchNorm's performance degrades severely when switching from minibatch to population statistics \citep{wu2021rethinking}.

We highlight the ability of normalization layers to act as spatial communication channels, providing a new perspective on such failures modes.
Previous work has shown that Convolutional Neural Networks (CNNs) learn position-aware representations around image boundaries via their receptive field \citep{islam2020much,kayhan2020translation}.
We further show that normalization-enabled networks learn a multi-hop message passing algorithm over the spatial dimension.
This allows CNNs to bypass translational equivariance and use information from beyond the local receptive field.

\emph{Our findings suggest care in the use of normalization layers in applications where preserving a limited receptive field is crucial.}
We consider as a case study the seminal work of \citet{janner2022diffuser} on diffusion models for robotic planning, which has been widely used as a foundation for subsequent work
\citep{ajay2022conditional,chen2023playfusion,chi2023diffusion,ze20243d,chen2025learning,chen2025extendable}.
Their method trains a diffusion model over expert trajectories with a one-dimensional convolutional denoising U-Net.
Here, the ``spatial'' dimension is in fact the trajectory's time dimension, and the authors emphasize that the \emph{temporal compositionality} enabled by the U-Net's limited receptive field is critical for flexibly stitching together expert subtrajectories during generation.
We conclusively demonstrate that the GroupNorm layer employed in the U-Net of \citet{janner2022diffuser} can in fact propagate information beyond the local receptive field, undermining temporal compositionality.

\inlinesubsection{Notation}
Let $\R$ denote the reals and $\N = \{1, 2, \ldots\}$ denote the natural numbers.
We define $[i] = \{1, 2, \ldots, i\}$ and $[i \isep j] = \{i, i+1, \ldots, j\}$ for $i, j \in \N$.
Set subtraction is written as $A \setminus B$ and cardinality as $|A|$.
We denote vectors $\bx \in \R^n$ using boldface.
Let $\E$ denote expectation and $\normalchar(\zeros{n}, \eye{n})$ be the standard normal distribution on $\R^n$.
For a measure $\mu$ and function $f$, we define $f \ \sharp \ \mu$ be the standard measure-theoretic pushforward of $\mu$ by $f$.
\section{The localization problem} \label{sec: problem}

We analyze a one-dimensional convolutional neural network $f_{\theta}: \Rh \to \Rh$, where $\h$ is the input sequence length:
\begin{align}
    \bz^{(1)} &= C_{in}(\bx), \nonumber \\
    \bz^{(i)} &= \sigma \big(\text{Norm}\big(C_i \big(\bz^{(i-1)}\big)\big)\big) \quad \forall i \in [2 \isep d], \label{eq: network_structure} \\
    f_{\theta}(\bx) &= C_{out}\big(\bz^{(d-1)}\big), \nonumber 
\end{align}
where $C_{in}, C_i, C_{out}$ are $1$-dimensional convolutional layers with parameters captured in $\theta$, $\text{Norm}$ is a normalization layer, and $\sigma$ is a ReLU nonlinearity. All convolutional layers are bias-enabled with an odd kernel size of $k$, zero-padding of width $\floor{\frac{k}{2}}$, unit stride, and zero dilation. Intermediate convolutions $C_i$ have an input-output hidden channel dimension $h \to h$, and $C_{in}$ and $C_{out}$ have input-output channel dimensions $1 \to h$ and $h \to 1$ respectively.

Let $\Fseq$ be the class of all functions $f: \Rh \to \Rh$ and $f_i: \Rh \to \R$ denote the function which picks out the $i$-th element of the output of $f$.
We derive the one-sided immediate receptive field of a depth-$d$ CNN to be $R(d) = \floor{\frac{k}{2}} \cdot d$, where the kernel size $k$ is an odd integer. For an index $i \in [1 + R(d) \isep \h - R(d)]$, $R(d)$ intuitively captures the largest shift $\Delta i$ such that $\frac{\partial f_i}{\partial x_{i \pm \Delta i}} (\bx) \neq 0$
for some normalization-free $f$ of the form \cref{eq: network_structure} and $\bx \in \Rh$.

\theoremspace
\begin{problem}[The $\F$-localization learning problem] \label{problem: localization}
The $\F$-localization learning problem, for some subclass $\F \subseteq \Fseq$, consists of learning a mapping $f \in \F$ which maps normally distributed inputs to a linearly increasing sequence by solving the optimization problem:
\begin{align*}
    \argmin_{f \in \F} \E_{\bx \sim \normalchar(\zeros{\h}, \eye{\h})} \| f(\bx) - \by \|_2^2,
\end{align*}
where the constant target $\by$ is defined as
\begin{align*}
    \by \defeq \left[ 0, \frac{1}{\h-1}, \frac{2}{\h-1}, \dots, \frac{\h-2}{\h-1}, 1 \right]  - \frac{1}{2} \in \Rh.
\end{align*}
\end{problem}

\Cref{problem: localization} is clearly trivial with an appropriate architecture \citep{liu2018intriguing}.
Conversely, it can only be partially solved by a normalization-free CNN with a limited receptive field $R < \h / 2$. Such networks can localize sequence indices where the receptive field overlaps the trajectory boundaries; but for indices far from the trajectory boundaries, the receptive field contains pure noise and is completely uninformative for localization.
\emph{We surprisingly show that the mere addition of normalization layers enables CNNs to solve this task by implicitly learning a multi-hop message passing algorithm along the normalization channels.}

\theoremspace
\begin{remark}
    The random inputs in \Cref{problem: localization} are necessary for information propagation, as we discuss in \Cref{sec: receptive_field_hypothesis}. Our setup most directly mirrors the use of convolutional U-Nets in diffusion models, where the input is corrupted by Gaussian noise \citep{janner2022diffuser}.
\end{remark}

\subsection{Detecting localization}
The previous discussion suggests that some functions $f \in \Fseq$, trained to solve \cref{problem: localization}, will only output ``correct locations'' for certain subsequences within $\by$. We now formalize what it means for a function $f \in \Fseq$ to be able to localize \emph{specific} sequence indices $i \in [1 \isep \h]$.


To this end, we define $P_{f_i}$ as the marginal distribution of predictions $f_i(\bx)$, under the input distribution $\bx \sim \normalchar(\zeros{\h}, \eye{\h})$.
Note that $P_{f_i}$ is a measure on $\R$ and is simply the pushforward measure of $\normalchar(\zeros{\h}, \eye{\h})$ under the function $f_i$:
\begin{align} \label{eq: marginal_distribution}
    P_{f_i} \defeq f_i \ \sharp \  \normalchar(\zeros{\h}, \eye{\h}).
\end{align}

Note that a convincing detection of $f$'s ability to localize a particular index $i^* \in [1 \isep \h]$ requires access to more than just $P_{f_{i^*}}$.
To see why, let $f(\bx) = \by_{i^*} \cdot \ones{\h}$ be a constant function.
In this case, $P_{f_{i^*}}$ is a point mass at $\by_{i^*}$, and appears to be ``correct'' for this index.
But evidently this prediction is not due to $f$'s ability to localize $i^*$ in the sequence.

We instead propose detecting $f$'s ability to localize $i^*$ by comparing $P_{f_{i^*}}$ to $P_{f_j}$ for $j \in [1 \isep \h] \setminus \{i^*\}$.

\theoremspace
\begin{definition}[Localization] \label{def: localization}
    We say that $f \in \Fseq$ \emph{localizes} an index $i^* \in [1 \isep \h]$ if
    \begin{align*}
        \E_{X_{i^*} \sim P_{f_{i^*}}} [X_{i^*}] \neq \E_{X_{j} \sim P_{f_{j}}} [X_{j}] \quad \forall j \in [1 \isep \h] \setminus \{i^*\},
    \end{align*}
    where the distributions $P$ are defined as in \cref{eq: marginal_distribution}.
\end{definition}

\Cref{def: localization} is made concrete by applying Welch's t-test \citep{welch1947generalization}. We also repurpose \Cref{def: localization} to discuss localization ability at an intermediate layer of a pre-trained network by simply grafting a probe onto the relevant activations, as detailed in \Cref{sec: experiments}.
\Cref{sec: non_localizable} theoretically shows that normalization-free CNNs cannot localize.

\theoremspace
\begin{remark}
    One limitation of \Cref{def: localization} is that it does not measure absolute correctness of $f$'s predictions; it only detects whether the network has learned to predict different positions for different sequence indices. We implicitly assume that these goals are aligned by the objective of \Cref{problem: localization}, and empirically find \Cref{def: localization} to be a satisfactory tool for detecting localization.
\end{remark}
\begin{figure*}[t]
    \centering
    \captionsetup[subfigure]{aboveskip=-0.0cm,belowskip=-0.1cm}
    \begin{subfigure}[t]{0.49\textwidth}
        \centering
        \includegraphics[width=\textwidth]{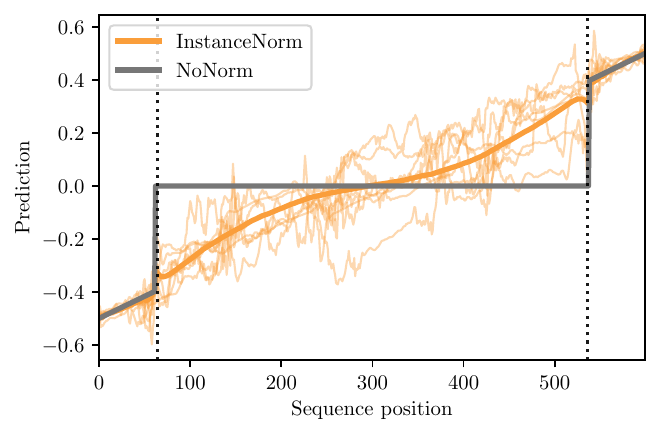}
        \subcaption{}
        \label{fig:illustrative_comparison_preds}
    \end{subfigure}
    \hfill
    \begin{subfigure}[t]{0.49\textwidth}
        \centering
        \includegraphics[width=\textwidth]{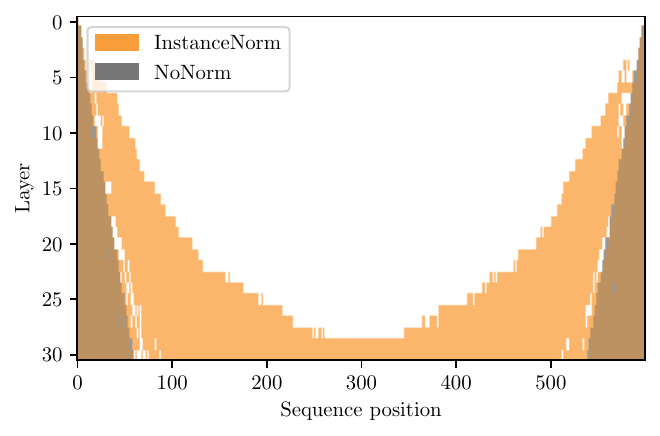}
        \subcaption{}
        \label{fig:illustrative_comparison_localization}
    \end{subfigure}
    \caption{
        \label{fig:illustrative_comparison_results}
    Localization ability for CNNs with and without normalization.
    (a) CNN predictions. Transparent thin lines are individual predictions, and the corresponding thick lines reflect the mean. The output sequences without normalization are near-identical under the plotted mean. Vertical dashed lines are located $R(d) = 64$ and $\h - R(d) = 536$ for reference.
    (b) Localization as a function of depth with and without InstanceNorm.
    A shaded cell indicates that a probe at that particular network depth was able to localize a particular sequence index as per \Cref{def: localization}.
    \vspace*{-0.3cm}
    }
\end{figure*}

\section{Experiments} \label{sec: experiments}
We show that normalization layers enable communication along the input's spatial dimension (\Cref{sec: normalization_communication}), present a hypothesized mechanism (\Cref{sec: receptive_field_hypothesis}), and show robustness to the choice of normalization (\Cref{sec: groupnorm_batchnorm}).
Experimental details are deferred to \Cref{sec: experimental_details}.

\subsection{Normalization layers are communication channels} \label{sec: normalization_communication}

\Cref{fig:illustrative_comparison_results} demonstrates how normalization layers enable localization for indices far from the sequence boundaries. We train two near-identical networks as described in \Cref{sec: problem}, one with InstanceNorm layers and one without.

\Cref{fig:illustrative_comparison_preds} shows the predictions of the CNN without normalization in gray. The receptive field of the network captures the convolutional zero-padding at the edges of the sequence for indices less than $R(d)$ and greater than $\h - R(d)$ (vertical dashed lines), leading to successful localization. Indices in the middle of the sequence are unable to be localized, and the predicted positions degenerate to the sequence mean of zero.
Conversely, the network with InstanceNorm layers localizes the entire sequence, as shown in orange.
Since input sequences are normally distributed, individual output predictions show some variability. However, the mean of the output predictions is clearly monotonically increasing.

\Cref{fig:illustrative_comparison_localization} inspects the propagation of localization information at intermediate network layers, as detected by training probes and applying \Cref{def: localization}.
For the network without normalization, location information propagates linearly from the edges of the sequence as expected.
However, once normalization layers are added, information propagates faster with depth and achieves complete localization by the last few layers.
We will show that this aligns with our hypothesized communication model, whereby the normalization layer can be used to communicate relative positional information for indices whose receptive fields overlap (\Cref{sec: receptive_field_hypothesis}). Deeper layers have larger receptive fields, and so can communicate information over longer distances.


\subsection{Overlapping receptive fields enable communication} \label{sec: receptive_field_hypothesis}

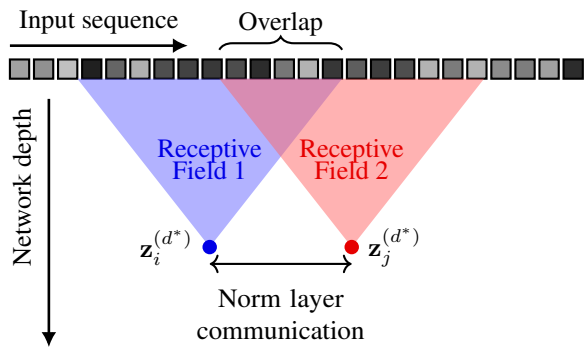
\begin{figure}[t]
    \centering
    \begin{tikzpicture}[scale=0.32]  
    \pgfmathsetseed{42}
    
    \def\coneHeight{7}
    \def\coneWidth{11}
    \def\leftConeX{8.3}
    \def\rightConeX{14.2}
    
    \pgfmathsetmacro{\overlapStart}{max(\leftConeX-\coneWidth/2, \rightConeX-\coneWidth/2)}
    \pgfmathsetmacro{\overlapEnd}{min(\leftConeX+\coneWidth/2, \rightConeX+\coneWidth/2)}
    
    \foreach \i in {0,...,23} {
        \pgfmathsetmacro{\shade}{20 + 70*rnd}
        \draw[fill=black!\shade!white, draw=black, thick] (\i*1.0,0) rectangle ++(0.8,0.8);
    }
    
    \fill[blue!60!white, opacity=0.5] 
        (\leftConeX-\coneWidth/2,0) -- 
        (\leftConeX+\coneWidth/2,0) -- 
        (\leftConeX,-\coneHeight) -- 
        cycle;
    
    \fill[red!60!white, opacity=0.5] 
        (\rightConeX-\coneWidth/2,0) -- 
        (\rightConeX+\coneWidth/2,0) -- 
        (\rightConeX,-\coneHeight) -- 
        cycle;
    
    \node[text=blue!90!black] at (\leftConeX, -3) {Receptive};
    \node[text=blue!90!black] at (\leftConeX, -3.8) {Field 1};
    
    \node[text=red!90!black] at (\rightConeX, -3) {Receptive};
    \node[text=red!90!black] at (\rightConeX, -3.8) {Field 2};
    
    \draw[thick, decorate, decoration={brace, amplitude=5pt}] 
        (\overlapStart, 1.2) -- (\overlapEnd, 1.2);
    
    \node[] at ({(\overlapStart+\overlapEnd)/2}, 2.3) {Overlap};
    
    \node[anchor=south west] (input_label) at (0.0, 1.5) {Input sequence};
    
    \draw[-{Latex[length=2mm, width=2mm]}, line width=1pt] 
        ([yshift=-1cm]input_label.west) -- ([yshift=-1cm]input_label.east);
    
    \node[rotate=90, anchor=south] (depth_label) at (1.5, -4.5) {Network depth};
    
    \draw[-{Latex[length=2mm, width=2mm]}, line width=1pt] 
        ([xshift=1cm]depth_label.east) -- ([xshift=1cm, yshift=-3cm]depth_label.west);
        
    \fill[blue!90!black] (\leftConeX, -\coneHeight) circle (0.3);
    \fill[red!90!black] (\rightConeX, -\coneHeight) circle (0.3);
    
    \node[anchor=east] at (\leftConeX-0.3, -\coneHeight) {$\mathbf{z}_i^{(d^*)}$};
    \node[anchor=west] at (\rightConeX+0.3, -\coneHeight) {$\mathbf{z}_j^{(d^*)}$};
    
    \draw[<->, thick, black] 
        (\leftConeX, -\coneHeight-0.7) -- (\rightConeX, -\coneHeight-0.7);
        
    \node[text width=4cm, align=center, text=black] 
        at ({(\leftConeX+\rightConeX)/2}, -\coneHeight-2.8) {Norm layer \\communication};
\end{tikzpicture}
    \caption{Illustration of the hypothesized mechanism for spatial communication through the normalization layer.}
    \vspace*{-0.4cm}
    \label{fig:mechanism}
\end{figure}

\Cref{sec: normalization_communication} shows that normalization layers enable communication along the input's spatial (sequence) dimension.
The precise mechanism is nonobvious.
Evidently, a fluctuation in the convolutional output at one sequence index can modulate other indices' activations since the normalization computations are spatially pooled; our perspective is that this is a crude communication channel which is broadcasted across the entire sequence.
But how can positional information transmission be \emph{localized} to neighboring indices in the sequence, resulting in the gradual propagation of positional information that we see in \Cref{fig:illustrative_comparison_localization}?
We now provide a potential mechanism for this communication.

We hypothesize that positional information is communicated through the normalization layer by exploiting the mutual information contained in overlapping receptive fields.
\Cref{fig:mechanism} illustrates this hypothesis.
Consider an intermediate network depth $d^*$ and sequence indices $i < j$ such that $i + R(d^*) > j - R(d^*)$.
As the blue receptive field cone rooted at $i$ captures part of the input sequence observed by the receptive field rooted at $j$, the activation $\bz^{(d^*)}_i$ potentially contains information about $\bz^{(d^*)}_j$.
This allows the network at index $i$ to then specifically transmit information to index $j$ through the normalization layer by modulating the mean and standard deviation of the sequence.

Note that this mechanism requires that the overlapping receptive field's pattern is somewhat unique in the input sequence.
The Gaussian noise used in \Cref{problem: localization} satisfies this for sufficiently long subsequences.
Uninformative input sequences, such as constant sequences, lead to no communication even when normalization layers are present.

To investigate this mechanism empirically, we sample Gaussian random input sequences $\bx, \tilde{\bx} \in \R^{\ell}$, where $\ell = 1 + 2R(d)$ and $d$ is the depth of the network.
We synthetically recreate an overlap $o \in \N$ between the sequences by setting $\tilde{\bx}_i \defeq \bx_{\ell - o + i}$ for $i \in [1 \isep o]$.
We define a modified CNN $f: \R^{\ell} \times \R^{\ell} \to \R^2$ as follows:
\begin{alignat}{10}
    \bz^{(1)} &= C_{in}(\bx), \; \tilde{\bz}^{(1)} = C_{in}(\tilde{\bx}), \nonumber \\
    \bu^{(i)} &= C_i\big(\bz^{(i-1)}\big), \; \tilde{\bu}^{(i)} = C_i\big(\tilde{\bz}^{(i-1)}\big) 
        \;\; &\forall i \in [2 \isep d], \nonumber \\
    \bv^{(i)}, \tilde{\bv}^{(i)} &= \text{PackNorm}\big(\bu^{(i)}, \tilde{\bu}^{(i)}\big) 
        &\forall i \in [2 \isep d], \nonumber \\
    \bz^{(i)} &= \sigma\big(\bv^{(i)}\big), \; \tilde{\bz}^{(i)} = \sigma\big(\tilde{\bv}^{(i)}\big) 
        &\forall i \in [2 \isep d], \nonumber \\
    f(\bx, \tilde{\bx}) &= \big[C_{out}\big(\bz^{(d-1)}\big), C_{out}\big(\tilde{\bz}^{(d-1)}\big)\big]. \nonumber
\end{alignat}
This is similar to the original network structure in \eqref{eq: network_structure} with the following key distinctions:
\begin{itemize}
    \item The network processes the two input sequences separately (but with shared weights).
    \item The only point of coupling between the two computational paths is PackNorm normalization layer, which concatenates $\bu^{(i)}$ and $\tilde{\bu}^{(i)}$ along the sequence dimension, applies InstanceNorm, and splits the result.
    \item All convolutions have no padding, resulting in the input sequences of length $\ell = 1 + 2R(d)$ ultimately mapping to output sequences of length $1$.
\end{itemize}

Our training objective is to regress $f(\bx, \tilde{\bx})$ onto the target $\by \defeq [-1, 1]$.
For zero overlap, this task is impossible as the output of $f$ would be equivariant to exchanging $\bx$ and $\tilde{\bx}$.
Our hypothesis predicts that as the overlap $o$ increases, the network should learn to transmit information via the shared PackNorm layer and correctly identify the sequences.

\begin{figure}[t]
    \centering
    \includegraphics[width=\columnwidth]{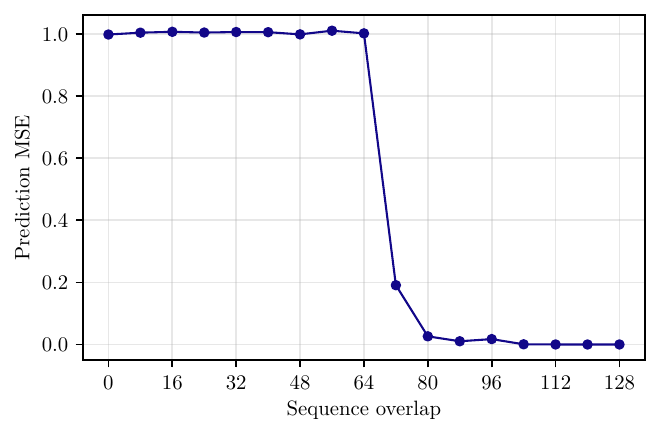}
    \vspace*{-0.5cm}
    \caption{
        Ability to distinguish sequences with overlapping receptive fields. A lower MSE loss indicates success.
    }
    \vspace*{-0.8cm}
    \label{fig:transmission_sweep}
\end{figure}

\Cref{fig:transmission_sweep} verifies this hypothesis and shows that a significant amount of overlap between the two input sequences is required for the network to successfully distinguish between them. Specifically, at about half the sequence length $\ell=129$ the network transitions from an MSE loss of $1$ (the loss of predicting a constant $[0, 0]$) to a near-zero loss.

This model allows us to be precise when we say that normalization layers enable multi-hop message passing in CNNs.
Specifically, we mean that the mechanism described above must occur multiple times in order to enable localization more than $3 \cdot R(d)$ indices away from the sequence boundaries.
The approximate factor of $3$ arises from considering a single message passing step at the deepest layer of the network, when the receptive field is at its maximum.
At this depth, the receptive field at $i=R(d)$ just touches the boundary of the sequence as well as the left side of the receptive field at $j=3 \cdot R(d)$; these two receptive fields make contact at $2 \cdot R(d)$ by symmetry. It is thus theoretically possible for a positional signal to propagate to a depth of $3 \cdot R(d)$ through the above model in a single step. However, \Cref{sec: normalization_communication} demonstrates localization across an entire trajectory even though $3 \cdot R(d) = 192$ is significantly less than $\ell / 2 = 300$. We can thus conclude that the proposed mechanism must have been applied iteratively at multiple depths of the network.

\subsection{Localization with GroupNorm and BatchNorm} \label{sec: groupnorm_batchnorm}

The localization results reported for InstanceNorm also transfer to GroupNorm and BatchNorm, with details deferred to \Cref{sec: additional_experiments}. For GroupNorm, we find that localization improves with the number of groups, matching the intuition that each group provides a distinct mean and variance to modulate---i.e., an added communication channel.

Remarkably, we also find that BatchNorm-equipped networks can localize for a sufficiently small batch size of $4$ and horizon of $250$.
This ability is impressive since communication is entangled across the batch dimension.
Our results suggest an alternative explanation for why BatchNorm performance degrades for small batches when switching from training to evaluation mode \citep{wu2021rethinking}.
Previous work widely holds that this empirical fact arises from distributional changes between population and batch statistics \citep{wu2021rethinking, ioffe2017batch,singh2019evalnorm,yan2020towards};
our results show that BatchNorm can use minibatch statistics at training time as a spatial communication channel which is entirely eliminated when substituting population statistics.
Conversely, we suggest that the elimination of this performance gap at larger batch sizes is partially attributable to crosstalk along the batch dimension subduing the ability of BatchNorm to act as a communication channel at training time.

\bibliography{references}
\bibliographystyle{icml2025}

\appendix
\onecolumn

\section{Normalization layers} \label{sec: normalization_layers}
We formally define the following normalization layers: Batch Normalization (BatchNorm) \citep{ioffe2015batch}, Layer Normalization (LayerNorm) \citep{ba2016layer}, Group Normalization (GroupNorm) \citep{wu2018group}, and Instance Normalization (InstanceNorm) \citep{ulyanov2016instance}.
We generally follow the notation in \citet{wu2018group} with some slight modifications.

\begin{definition}[Normalization Layer]
    \label{def:normalization}
    We consider an input $Z \in \R^{N \times C \times S}$ where $N, C, S$ are the batch axis, channel axis and spatial axis (e.g., height $\times$ width in 2D images).
    Inputs are indexed by a multi-index
    \[
        i \defeq (i_N, i_C, i_S) \in [N] \times [C] \times [S].
    \]
    The normalization layer computes the output as follows:
    \begin{align}
        \text{Norm}(Z)_i \defeq \gamma_{i_C} \frac{Z_i - \mu_i}{\sigma_i} + \beta_{i_C} \in \R^{N \times C \times S},
    \end{align}
    where $\mu$ and $\sigma$ are computed from $Z$, and $\gamma$ and $\beta$ are learnable parameters. The mean and standard deviation are computed as follows:
    \begin{align}
        \mu_i = \frac{1}{|\Sc_i|} \sum_{j \in \Sc_i} Z_j, \quad
        \sigma_i = \sqrt{\frac{1}{|\Sc_i|} \sum_{j \in \Sc_i} (Z_j - \mu_i)^2},
    \end{align}
    where $\Sc_i \subset [N] \times [C] \times [S]$ is the index set over which the mean and standard deviation are computed. The normalization layer can be one of the following:
    \begin{itemize}
        \item \textbf{BatchNorm}: Mean and standard deviation are computed over the batch and spatial dimensions:
        \[
            \Sc_i = \left\{(n, i_C, s) \mid n \in [N], s \in [S]\right\}.
        \]
        \item \textbf{LayerNorm}: The mean and standard deviation are computed over the channel and spatial dimensions:
        \[
            \Sc_i = \left\{(i_N, c, s) \mid c \in [C], s \in [S]\right\}.
        \]
        \item \textbf{InstanceNorm}: The mean and standard deviation are computed over the spatial dimensions:
        \[
            \Sc_i = \left\{(i_N, i_C, s) \mid s \in [S]\right\}.
        \]
        \item \textbf{GroupNorm}: The channels are partitioned into $G$ groups, each containing $K := C/G$ channels. Then the mean and standard deviation are computed over the group as well as the spatial dimensions:
        \[
            \Sc_i = \left\{(i_N, c, s) \mid c \in \big[K \floor{i_C / K} \isep K (\floor{i_C / K} + 1) - 1 \big], s \in [S]\right\}.
        \]
        Notably, GroupNorm recovers LayerNorm when $G = 1$ and InstanceNorm when $G = C$.
    \end{itemize}
\end{definition}

We note that BatchNorm maintains an Exponential Moving Average (EMA) of the mean and standard deviation which is used in evaluation mode.

\section{Related literature}
BatchNorm is a standard tool for accelerating training and improving performance on computer vision tasks, such as image classification \citep{he2016deep}, object detection \citep{he2017mask} and segmentation \citep{ronneberger2015u}.
However, it is well-known that BatchNorm performs poorly for small batch sizes \citep{wu2018group,wu2021rethinking}.
Previous works have connected this to BatchNorm's unique behavior of switching from minibatch to population statistics at evaluation time \citep{wu2021rethinking}.
Namely, authors have hypothesized that for small batch sizes, minibatch statistics distributionally differ from the population statistics computed via EMA \citep{wu2021rethinking,singh2019evalnorm,yan2020towards}.
This has even led to the suggestion of using minibatch statistics at evaluation time \citep{wu2021rethinking}.
Our work demonstrates that this explanation is incomplete; minibatch statistics can in fact provide a spatial communication channel which disappears when switching to population statistics.

LayerNorm, GroupNorm, and InstanceNorm have been widely used in generative models, from Generative Adversarial Networks (GANs) and diffusion models for images \citep{karras2019style,ho2020denoising} to trajectory generation \citep{janner2022diffuser}.
Early analysis on GANs has found that spatially pooled normalization layers can lead to blob-shaped artifacts by eliminating the relative magnitudes of different feature channels \citep{karras2020analyzing}.
The recent work of \citet{karras2024analyzing} further speculates that global normalization can negatively impact performance as it forces interactions between distant pixels.
\citet{kamb2024analytic} identifies that convolutional diffusion models without normalization layers can generate highly creative images by enforcing locality and equivariance.
We analyze normalization layers in a controlled setting and find for the first time that they can enable a sophisticated iterative communication scheme.
This property is at odds with generative applications where a limited receptive field is crucial \citep{janner2022diffuser,ajay2022conditional,chi2023diffusion}.

\section{Normalization-free CNNs cannot localize} \label{sec: non_localizable}

We have informally argued why normalization-free CNNs cannot localize.
Here we make this statement precise and include a brief proof sketch.
The following proposition is a trivial application of translation equivariance, and we include it to highlight the discriminatory power of \Cref{def: localization}.

\theoremspace
\begin{proposition} \label{prop: non_localizable}
    Let $f \in \Fseq$ be a normalization-free CNN of depth $d$ with a ReLU nonlinearity (see \Cref{eq: network_structure}). Assume $R(d) < \frac{\h}{2} - 2$. Then $f$ cannot localize any index 
    \begin{align} \label{eq: non_localizable}
        i^* \in \big[ R(d) + 1 \; \isep \; \h - R(d) - 1 \big]
    \end{align}
    in the sense of \Cref{def: localization}.
\end{proposition}
\vspace*{-0.2cm}
\begin{proof}[Proof (informal sketch)]
    By translation equivariance \citep{kondor2018generalization}, we know that $f_i = f_j$ for all $i, j$ in interval \eqref{eq: non_localizable}. Since $R(d) < \frac{\h}{2} - 2$, there exists a $j \neq i^*$ in \eqref{eq: non_localizable} such that $f_{i^*} = f_j$. Thus $P_{f_{i^*}} = P_{f_j}$ and $\E_{X_{i^*} \sim P_{f_{i^*}}} [X_{i^*}] = \E_{X_{j} \sim P_{f_{j}}} [X_{j}]$, violating \Cref{def: localization}.
\end{proof}

\section{Additional experiments} \label{sec: additional_experiments}

\begin{figure}[t]
    \centering
    \includegraphics[width=0.49\textwidth]{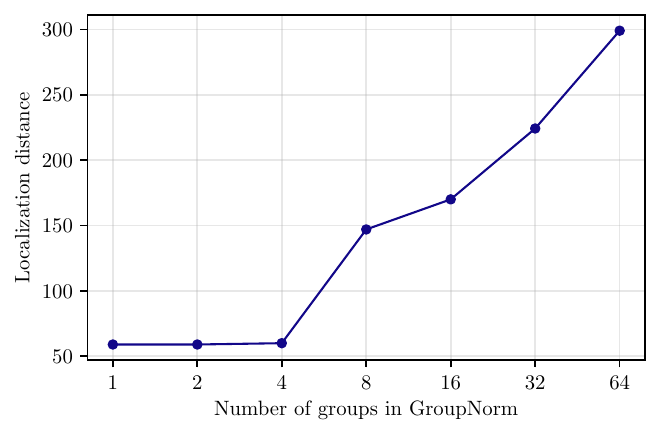}
    \caption{
        \label{fig:groupnorm_sweep}
        GroupNorm performance as a function of the number of groups.
        We plot the average localization distance over five distinct random seeds for each group count. The standard deviation is too small to be plotted.
    }
\end{figure}

\textbf{\Cref{sec: groupnorm_batchnorm}.} \Cref{fig:groupnorm_sweep} plots the localization distance of a GroupNorm-equipped CNN as a function of the number of groups. We define the ``localization distance'' as the largest distance from the edge of the sequence that can be localized as per \Cref{def: localization}, using a probe trained on the final layer for consistency with the localization plots (e.g. \Cref{fig:illustrative_comparison_localization}). Thus for a sequence length of $600$, a localization distance of $300$ indicates that the entire sequence has been localized by the final layer. We take our group sizes to be power-of-two divisors of the hidden dimension $64$ and consider a sequence length of $\h = 600$. Each group provides a distinct mean and variance which can be modulated along the sequence dimension. We find that localization distance increases as we add more groups, consistent with the intuition that each group provides an additional communication channel.

\begin{figure*}[t]
    \centering
    \captionsetup[subfigure]{aboveskip=-0.0cm,belowskip=-0.1cm}
    \begin{subfigure}[t]{0.49\textwidth}
        \centering
        \includegraphics[width=\textwidth]{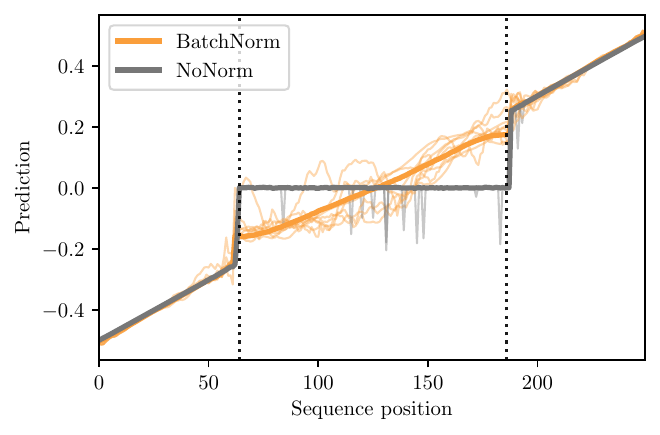}
        \subcaption{}
        \label{fig:batchnorm_illustrative_comparison_preds_BatchNorm_NoNorm}
    \end{subfigure}
    \hfill
    \begin{subfigure}[t]{0.49\textwidth}
        \centering
        \includegraphics[width=\textwidth]{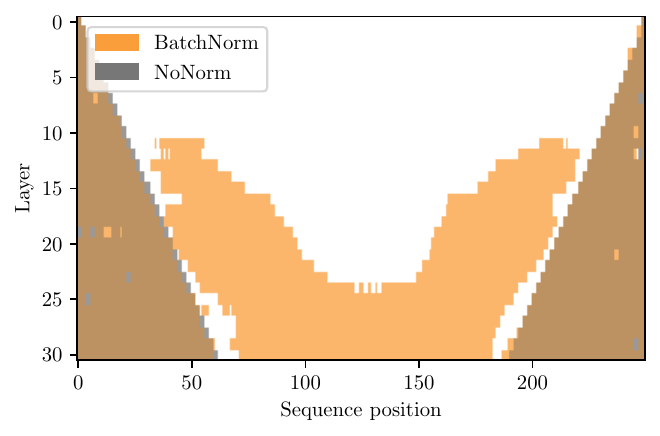}
        \subcaption{}
        \label{fig:batchnorm_illustrative_comparison_localization_BatchNorm_NoNorm}
    \end{subfigure}
    \caption{
        \label{fig:batchnorm_illustrative_comparison_results_BatchNorm_NoNorm}
    Localization ability for CNNs with and without BatchNorm layers (using minibatch statistics at inference time).
    (a) CNN predictions. Transparent thin lines are individual predictions, and the corresponding thick lines reflect the mean. Vertical dashed lines are located $R(d) = 64$ and $\h - R(d) = 186$ for reference.
    (b) Localization as a function of depth.
    A shaded cell indicates that a probe at that particular network depth was able to localize a particular sequence index as per \Cref{def: localization}.
    \vspace*{-0.3cm}
    }
\end{figure*}

\Cref{fig:batchnorm_illustrative_comparison_results_BatchNorm_NoNorm} demonstrates that BatchNorm is also able to act as a communication channel. We consider a shorter sequence length of $\h = 250$ and a batch size of $4$. \emph{Crucially, we run BatchNorm with minibatch statistics at inference time---substituting population statistics destroys localization ability.} We compare to an identical network without BatchNorm. Again, the normalization-free network is unable to localize central sequence indices. Localization with BatchNorm succeeds, although it is more challenging than GroupNorm due to communication crosstalk along the batch dimension.

We note that there appears to be some failure in localization with BatchNorm around the boundary of the normalization-free network's localized region. This corresponds to the indices whose receptive field just touches the boundary of the sequence. We observe anecdotally that network predictions can be quite noisy in this transitional region, meaning that more samples would need to be collected for the computation of Welch's t-test to reliably reject the null hypothesis.

\begin{figure*}[t]
    \centering
    \captionsetup[subfigure]{aboveskip=-0.0cm,belowskip=-0.1cm}
    \begin{subfigure}[t]{0.49\textwidth}
        \centering
        \includegraphics[width=\textwidth]{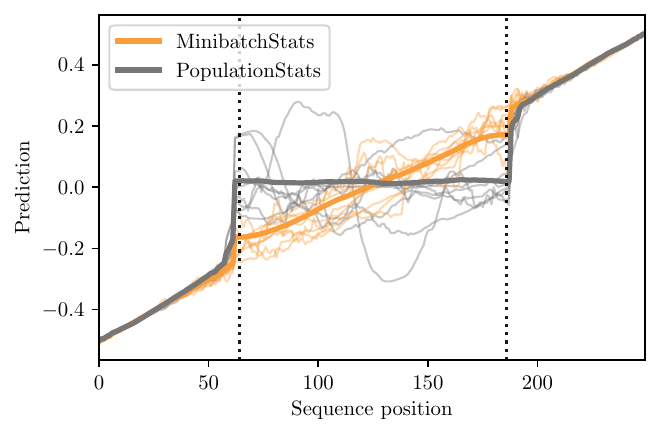}
        \subcaption{}
        \label{fig:batchnorm_illustrative_comparison_preds_MinibatchStats_PopulationStats}
    \end{subfigure}
    \hfill
    \begin{subfigure}[t]{0.49\textwidth}
        \centering
        \includegraphics[width=\textwidth]{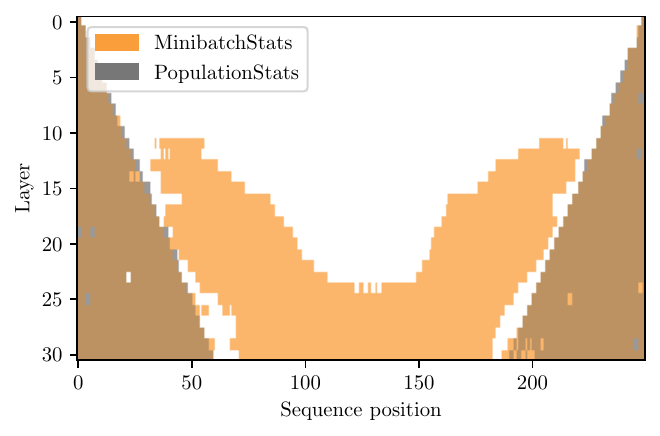}
        \subcaption{}
        \label{fig:batchnorm_illustrative_comparison_localization_MinibatchStats_PopulationStats}
    \end{subfigure}
    \caption{
        \label{fig:batchnorm_illustrative_comparison_results_MinibatchStats_PopulationStats}
    Localization ability for CNN with BatchNorm layers, using either minibatch or population statistics at inference time.
    (a) CNN predictions. Transparent thin lines are individual predictions, and the corresponding thick lines reflect the mean. Vertical dashed lines are located $R(d) = 64$ and $\h - R(d) = 186$ for reference.
    (b) Localization as a function of depth.
    A shaded cell indicates that a probe at that particular network depth was able to localize a particular sequence index as per \Cref{def: localization}.
    \vspace*{-0.3cm}
    }
\end{figure*}

\Cref{fig:batchnorm_illustrative_comparison_results_MinibatchStats_PopulationStats} shows that using population statistics at inference time eliminates the ability of BatchNorm to act as a communication channel.
This corroborates well-known results in the literature that switching BatchNorm to population statistics at inference time induces performance degradation for small batch sizes.
However, we find previous explanations for this phenomenon as arising from a change in activation distributions to be incomplete \citep{wu2021rethinking, ioffe2017batch,singh2019evalnorm,yan2020towards}. 
Our results show that BatchNorm can use minibatch statistics at training time as a spatial communication channel which is entirely eliminated when switching to population statistics.

\section{Experimental details} \label{sec: experimental_details}

\textbf{\Cref{sec: normalization_communication}.}
We train over sequences of length $\h = 600$ with Adam for $10^{5}$ iterations with a batch size of $32$ and learning rate of $10^{-4}$. Both trained CNNs have a depth of $32$ and kernel size of $5$, resulting in an immediate receptive field of size $64$. The hidden dimension is $64$.

We train probes for $5 \cdot 10^3$ iterations using Adam with a learning rate of $10^{-3}$ and a batch size of $32$ sequences. One probe $g_{\phi}^{(i)}: \R^h \to \R$ is trained over each hidden activation $\bz^{(i)} \in \R^{\ell \times h}$ for $i \in [1 \isep d]$ (see \Cref{eq: network_structure}). Specifically, we treat $\bz^{(i)}$ as a batch of $\ell$ samples and regress the $i$-th batch element on the target $\by_i$ using the MSE loss. Each probe is a simple ReLU-activated multi-layer perceptron with two hidden layers of width $256$.

We compute the statistical tests for \Cref{def: localization} using Welch's t-test with a p-value threshold of $0.1$ over $32 \times 10^{5}$ predictions.

\textbf{\Cref{sec: receptive_field_hypothesis}.}
Our model has depth $d=32$, a kernel size of $5$, and a hidden dimension of $64$. We train using Adam with a learning rate of $10^{-4}$ and a batch size of $32$ for $10^4$ iterations.

\textbf{\Cref{sec: groupnorm_batchnorm}.} All models have a depth of $d=32$, a kernel size of $5$, and a hidden dimension of $64$. We vary the number of groups $G$ for all GroupNorm layers, and report the best localization score over a learning rate sweep of $\{10^{-5} \cdot 4^i \mid i \in [0 \isep 3]\}$. We train using Adam with a batch size of $32$ for $10^5$ iterations. Probe and Welch's t-test parameters are the same as in \Cref{sec: normalization_communication}. 

For the BatchNorm experiments, we use a length $\h = 250$ and a batch size of $4$. Training is done with Adam with a learning rate of $10^{-4}$ for $5 \cdot 10^5$ iterations. Our networks have a depth of $32$, kernel size of $5$, and hidden dimension of $64$. We accumulate gradients for $8$ batches to maintain an effective batch size of $32$. Probe and Welch's t-test parameters are the same as in \Cref{sec: normalization_communication}.

\end{document}